\documentclass[acmtog, 9pt]{acmart}

\AtBeginDocument{%
  }

\usepackage{hyperref} 
\usepackage{graphicx}
\usepackage{textcomp}
\usepackage{xcolor}
\usepackage{algorithm}
\usepackage{algpseudocode}
\usepackage{subfig}

\newcommand{\D}{\mathcal{D}}
\newcommand{\X}{\mathcal{X}}
\newcommand{\Z}{\mathcal{Z}}

\newcommand{\LS}{\mathcal{L}}

\renewcommand*{\bibfont}{\footnotesize}

\copyrightyear{2024}
\acmYear{2024}
\setcopyright{rightsretained}
\acmConference[DAC '24]{61st ACM/IEEE Design Automation Conference}{June 23--27, 2024}{San Francisco, CA, USA}
\acmBooktitle{61st ACM/IEEE Design Automation Conference (DAC '24), June 23--27, 2024, San Francisco, CA, USA}
\acmDOI{10.1145/3649329.3656543}
\acmISBN{979-8-4007-0601-1/24/06}

\begin{document}

\title{CircuitVAE: Efficient and Scalable Latent Circuit Optimization}
\author{Jialin Song}
\authornote{Equal Contribution}
\email{jialins@nvidia.com}
\affiliation{ 
\institution{NVIDIA}
\country{USA}
}
\author{Aidan Swope}
\authornote{Work done while at NVIDIA}
\authornotemark[1]
\affiliation{ 
\institution{Harmonic}
\country{USA}
}
\author{Robert Kirby}
\affiliation{ 
\institution{NVIDIA}
\country{USA}
}
\author{Rajarshi Roy}
\affiliation{ 
\institution{NVIDIA}
\country{USA}
}
\author{Saad Godil}
\authornotemark[2]
\affiliation{ 
\institution{Hippocratic AI}
\country{USA}
}
\author{Jonathan Raiman}
\affiliation{ 
\institution{NVIDIA}
\country{USA}
}
\author{Bryan Catanzaro}
\affiliation{ 
\institution{NVIDIA}
\country{USA}
}

\begin{abstract}
    Automatically designing fast and space-efficient digital circuits is challenging because circuits are discrete, must exactly implement the desired logic, and are costly to simulate.
    We address these challenges with CircuitVAE, a search algorithm that embeds computation graphs in a continuous space and optimizes a learned surrogate of physical simulation by gradient descent.
    By carefully controlling overfitting of the simulation surrogate and ensuring diverse exploration, our algorithm is highly sample-efficient, yet gracefully scales to large problem instances and high sample budgets.
    We test CircuitVAE by designing binary adders across a large range of sizes, IO timing constraints, and sample budgets.
    Our method excels at designing large circuits, where other algorithms struggle: compared to reinforcement learning and genetic algorithms, CircuitVAE typically finds 64-bit adders which are smaller and faster using less than half the sample budget.
    We also find CircuitVAE can design state-of-the-art adders in a real-world chip, demonstrating that our method can outperform commercial tools in a realistic setting.
\end{abstract}

\maketitle

\section{Introduction}
\begin{figure*}[t]
    \centering
    \includegraphics[width=0.9\textwidth]{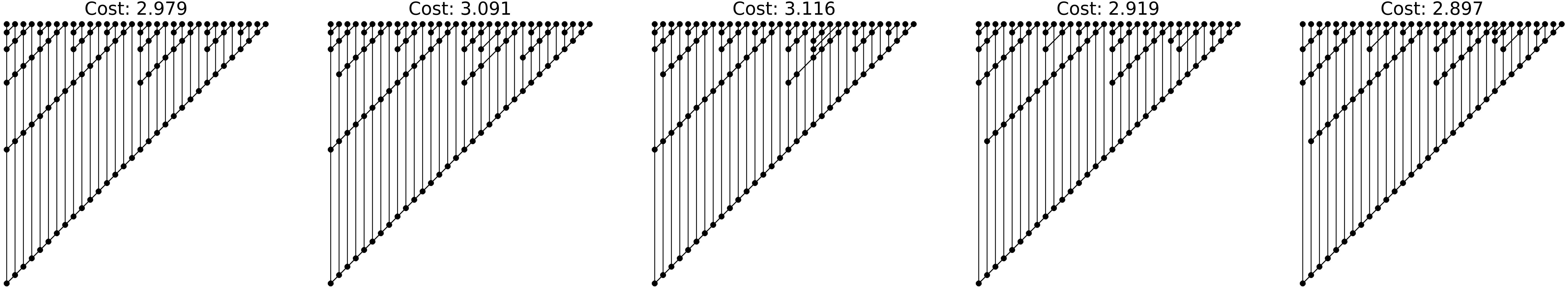}
    \caption{A sample evolution of 32-bit adders discovered by CircuitVAE. Starting from the Sklansky structure, CircuitVAE iteratively navigates a learned latent space to produce different designs until reaching a lowest cost one on the right.}
\label{fig:progression}
\vspace{-0.1in}
\end{figure*}
As the workhorses of today's parallel processors, optimizing the design of binary adders is an important and well-studied problem.
However, because the physical characteristics of a given design change as manufacturing technology improves, and because each adder in a larger design may face different constraints, classical adder designs which minimize analytical properties such as circuit depth often perform poorly in practice. 
More recent approaches use physical synthesis and simulation to optimize adders for real-world area, delay, and power consumption \cite{prefixrl, songmulti2022}. Such algorithms remain expensive, though, because physical synthesis is slow, the problem is discrete, and the search space grows exponentially with the number of bits to be summed. Therefore, optimizing larger adders has generally remained intractable.

We present CircuitVAE, a highly sample-efficient algorithm for optimizing binary adders which outperforms human designs and commercial tools while requiring fewer simulations than competing approaches.
CircuitVAE solves the two key challenges of this domain, discrete search and an expensive objective function, by embedding circuits in a continuous space and learning to predict the results of physical simulation.
We use a $\beta$-VAE \cite{higgins2017beta} to represent adders as continuous latent vectors, jointly trained with a neural cost predictor whose gradients help organize the latent space according to cost.
At search time, we use gradient descent through the cost model to search for latent vectors minimizing predicted cost.
We introduce two domain-agnostic improvements to the standard latent-space optimization framework. 
Our first, prior-regularized search, prevents search points from ``overfitting'' the cost predictor far from the data manifold.
The second, cost-weighted sampling, helps balance quality and diversity in the explored points by initializing them from high-performing prior evaluations.
Through extensive ablations, we demonstrate that these improvements enable gradient-based search to outperform Bayesian optimization in the latent space, contrary to the standard approach.

We evaluate CircuitVAE in numerous settings, both on standard benchmarks and in the real world.
Across various sizes of adders, and emulating various cost tradeoffs between area and delay, we find that CircuitVAE outperforms human designs, commercial tools, and the prior state-of-the-art reinforcement learning algorithm in cost and simulation requirements.
Finally, we integrate CircuitVAE into a real-world chip design workflow and show that it outperforms commercial tools in a realistic setting.

\section{Related work}
\subsection{Machine learning for EDA}
The most closely related work to ours is  \cite{prefixrl}, which also optimizes parallel prefix adders with deep learning.
While their reinforcement learning (RL) approach outperforms conventional tools, we show that RL is hindered by the difficulty of searching directly in the input space.
In head-to-head comparison (\autoref{subsec:comparison}), CircuitVAE is typically more than twice as data-efficient as RL due to learning its own well-structured search space.
Classical approaches to this problem include heuristic search \cite{moto2018prefix} and pruning \cite{roy2014towards} methods.

Several works have explored using machine learning for other parts of the electronic design automation (EDA) process.
These include \cite{googlePlace} and \cite{Lin2020DREAMPlaceDL}, who use deep learning for chip placement, \cite{wang2020learning} who use RL to design analog rather than digital circuits, and \cite{zuo2023rl} who use RL to design multipliers.

We build on the open-source EDA tools OpenROAD \cite{ajayi2019openroad} and OpenPhySyn \cite{Agiza2020OpenPhySynAO}, without which this work would not have been possible.

\subsection{Latent-space optimization}
CircuitVAE employs latent-space optimization (LSO), a method which has recently become popular for black-box optimization, most notably in the field of chemical design \cite{chemvae, jin2018junction, tripp2020sample}.
LSO consists of learning a latent-variable generative model over input structures, together with a neural predictor of the cost function which serves to shape the latent space and may be used for optimization.
The latent space acts as a learned continuous search space, typically grouping semantically similar inputs and enabling continuous search algorithms.
While many improvements to this scheme have been recently proposed, we build on the framework of \cite{tripp2020sample}, which interleaves optimization with retraining the generative model on new data.

While some LSO techniques optimize by gradient descent through the cost predictor, recently Bayesian optimization (BO) has been the preferred approach \cite{jin2018junction}.
In this work, we demonstrate that naive gradient descent suffers from over-optimizing the cost predictor far from the data manifold, yielding points with low predicted costs but high actual costs.
However, we introduce two techniques to address this problem and show that, once appropriately regularized, gradient descent can outperform BO by a wide margin.

\section{Background}
\label{subsec:background}

Many kinds of circuits, notably including binary adders, can be compactly represented as \emph{prefix graphs}, which describe the pattern of carries within the circuit.
Prefix graphs compactly represent a circuit's design in terms of carry \emph{generation} and \emph{propagation} \cite{bk}.
Each bit span \texttt{i:j} is associated with a generate bit $g_{i,j}$ and a propagate bit $p_{i,j}$.
For all $i=1 \ldots N$, computing the input bits $g_{i,i}$ and $p_{i,i}$ is straightforward; furthermore, given the output bits $(g_{1,1}; p_{1,1}) \ldots (g_{N,1}, p_{N,1})$, computing the final carries and summand is easy.
Intermediate values may be computed recursively: $(g_{i, j}; p_{i,j}) = (g_{i, x}; p_{i,x}) \circ (g_{x-1, j}; p_{x-1,j})$ where $i \geq x > j$ and $\circ$ is the carry operator Brent and Kung describe.
A prefix graph is exactly a tree determining the association order of $(g_{i,i}; p_{i,i}) \circ \ldots \circ (g_{1,1}; p_{1,1})$ for each $i = 1, \ldots, N$.

The design space of prefix graphs is large---$O(2^{N^2})$ for $N$-bit designs---and expresses tradeoffs between circuit area and delay, the two main desiderata.
For example, the ripple-carry structure represents schoolbook addition, computing carries one bit at a time, and has the lowest possible area but is relatively slow.
Faster adders such as Sklansky \cite{sklansky1960conditional} and Kogge-Stone \cite{kogge1973a} compute redundant carry bits, which enables some parallelism at the cost of area.
While regular structures minimizing analytical properties like graph depth and connectivity are well-known, the actual delay of a fully synthesized and laid-out circuit depends in a complicated way on many other physical factors, so designing these circuits remains an important industrial problem.

Because real-world designs may have different requirements for area and latency, we define a scalar cost function $f(x) = \omega \cdot \text{delay}(x) + (1 - \omega) \cdot \text{area}(x)$.
We call $\omega$ the \emph{delay weight}, a hyperparameter trading off these competing goals.
In the cost function, we measured a circuit's total area in square microns divided by 100 and the delay of its longest (``critical'') path in nanoseconds multiplied by 10, as we found this yielded smooth changes in optimization as $\omega$ was swept from 0 to 1. 
To ensure our results generalize, we conducted our experiments for $\omega \in \{0.33, 0.66, 0.95\}$ and for 32-, and 64-bit adders, as well as an experiment for designing 26-bit gray-to-binary converters with $\omega = 0.6$.

Prefix graphs are translated into physical circuits through cell mapping (which translates the logical graph into a list of electrical components with a lookup table), physical synthesis, and layout.
In this work we experiment with binary adders and gray-to-binary converters \cite{doran2007gray}, but the algorithm we describe could straightforwardly apply to any parallel prefix circuit by altering the cell mapping process.

\section{CircuitVAE}
\begin{figure}[t]
    \centering
    \includegraphics[width=0.5\textwidth]{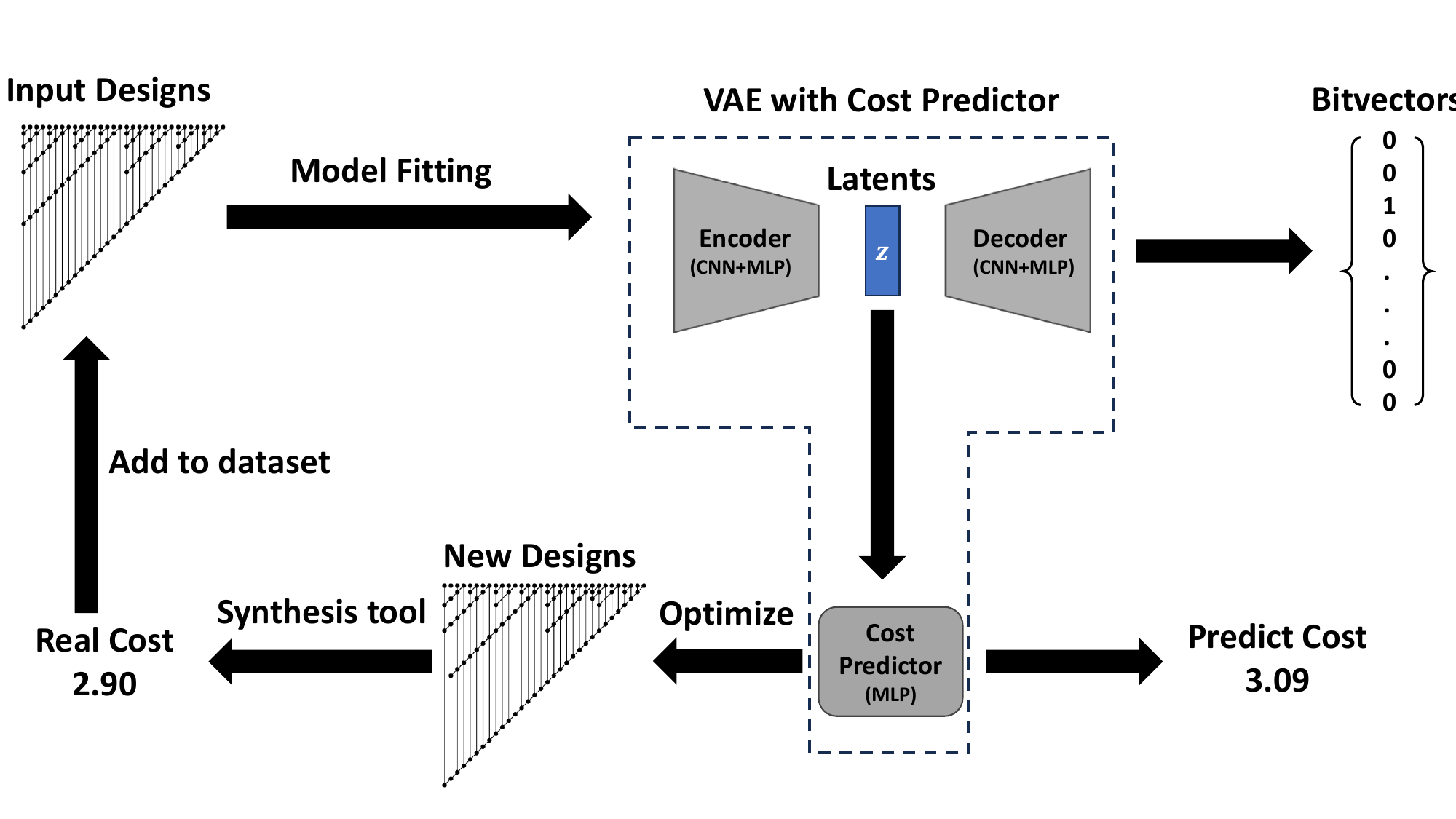}
    \caption{The CircuitVAE algorithm flowchart. We first fit a VAE equipped with a cost predictor on an input dataset. We use prior-regularized search (\autoref{method:optimization}) to optimize the cost predictor to generate new designs. We repeat this loop multiple times to gradually discover better designs.}
\label{fig:method}
\vspace{-0.1in}
\end{figure}
In this section, we describe our CircuitVAE algorithm for optimizing prefix adders. \autoref{fig:method} gives an overview of the algorithm.
In \autoref{method:training}, we describe how to train CircuitVAE by combining the standard VAE training objective with a cost predictor loss.
In \autoref{method:optimization}, we describe how to search in the latent space of CircuitVAE, including our proposed prior-regularized search and cost-weighted sampling techniques. 

\subsection{Training}
\label{method:training}
We denote by $\X$ the discrete search space for all $N$-bit adders.  
Optimizing over $\X$ is challenging: computation graphs with many nodes or edges in common may not have similar costs because removing one node is sufficient to change the critical path.
Therefore, we embed $\X$ into a continuous latent space $\mathcal{Z}$ with a VAE augmented with a cost prediction head.

Concretely, we learn an encoding function $q_{\phi}(z|x)$ that encodes an input $x$ to a distribution over $\Z$, and a decoding function $p_{\theta}(x|z)$ that decodes a latent variable $z$ to a distribution over $\X$. 
We optimize the parameters $\theta$ and $\phi$ by maximizing the evidence lower-bound (ELBO) \cite{kingma2013auto}: $\mathbb{E}_{q_{\phi}(z|x)}[\log p_{\theta}(x|z)] - D_{KL}(q_{\phi}(z|x) || p_{\theta}(z))$, where $D_{KL}$ is the Kullback–Leibler divergence between two distributions.
Our prior $p_{\theta}(z)$ is a diagonal unit Gaussian.
To balance adherence to the prior with other training objectives, we use a $\beta$-VAE \cite{higgins2017beta, bozkurt2021rate}.
Our training loss is:
\begin{equation}
    \LS_{\theta, \phi}(x) = \mathbb{E}_{q_{\phi}(z|x)}[\log p_{\theta}(x|z)] - \beta D_{KL}(q_{\phi}(z|x) || p_{\theta}(z))
\end{equation}

Given a dataset of prefix adders and their costs $\D = \{(x_i, c_i)\}_{i=1}^n$, we can fit the VAE parameters by maximizing $\sum_{i=1}^n \LS_{\theta, \phi}(x_i)$ via gradient descent using the reparameterization trick \cite{kingma2013auto}. 

In addition to learning the encoder and decoder, we also learn a cost predictor model $f_{\pi}: \Z \rightarrow \mathbb{R}$.
For a datapoint $(x, c)$, we first map $x$ to a latent space distribution $q_{\phi}(z|x)$, next we sample a latent variable $z \sim q_{\phi}(z|x)$ (again using the reparametrization trick), and finally we predict its cost $f_{\pi}(z)$.
Our cost prediction loss is $\LS_{\pi}(x, c) = (f_{\pi}(z) - c)^2$. 
The cost predictor enables optimization, but it also helps shape the latent space: observe that if two circuits $x_1, x_2$ with very different costs have overlapping posteriors $q_{\phi}(\cdot \mid x_1), q_\phi(\cdot \mid x_2)$, the cost predictor will fail to distinguish them.
Therefore, the training loss is minimized when circuits with similar costs are grouped together, which aids optimization.

Following \cite{tripp2020sample}, we reweight datapoints according to their cost to give promising points more volume in latent space.
Specifically, the weight of a datapoint $(x, c) \in \D$ is
\begin{equation}
\label{eq:weighting_function}
    w(x; \D, k) \propto \frac{1}{kn + \text{rank}_{\D}(x)}, \quad \newline 
    \text{rank}_{\D}(x) = \left|\left\{x_i : c_i < c,\ (x_i, c_i) \in \D \right\}\right|\ ,
\end{equation}
where $k$ is a hyperparameter controlling the relative weights among the datapoints.
For simplicity, we use $w_i(\D)$ to denote the normalized weight for a datapoint $(x_i, c_i)$. Note that we need to recompute these weights after acquiring new datapoints because of the dependency on $\D$.

Finally, we can put the two losses together to obtain the overall loss objective 
\begin{equation}
\label{eq:vae_loss}
    \LS_{\theta, \phi, \pi}(\D) = \sum_{i=1}^n w_i(\D)\LS_{\theta, \phi}(x_i) + \lambda \LS_{\pi}(x_i, c_i)
\end{equation}
where $\lambda \in \mathbb{R}^{+}$ is a hyperparameter to balance the VAE training loss and the cost prediction loss.
In all experiments, we set $\beta = 0.01$, $\lambda = 10.0$, and $k = 0.001$, and optimize $\LS_{\theta, \phi, \pi}$ with Adam \cite{kingma2014adam}.

\begin{algorithm}
\caption{CircuitVAE}\label{alg:circuitvae}
\begin{algorithmic}[1]
\State {\bf Input}: $\D_0$ initial dataset; $f$ a blackbox function available for queries; $M$ the number of data acquisition rounds; $m$ the number of parallel latent search; $T$ the number of gradient descent steps for latent space optimization; $t$ the interval to capture latents during optimization

\State $\D \gets \D_0$
\For{$i=1...M$} 
\State Compute sample weights for $\D$ (Equation \ref{eq:weighting_function})
\State Fit parameters $(\theta, \phi, \pi)$ of VAE with the cost predictor on $\D$ with the weighted training objective (Equation \ref{eq:vae_loss})
\State Sample $m$ points from $\D$ proportional to sample weights
\State Sample $m$ initial latents with $q_{\phi}$
\State Optimize $g(z)$ (Equation \ref{eq:latent_search}) with gradient descent from the initial latents for $T$ steps and capture a set of latents $Z_i$ along the optimization trajectory after every $t$ gradient steps
\State Sample a new set of $X_i$ by decoding $Z_i$ through $p_{\theta}$
\State Query $f$ on $\X_t$ to obtain $\D_i$
\State $\D \gets \D \cup \D_i$
\EndFor
\State \Return the lowest cost point in $\D$
\end{algorithmic}
\end{algorithm}

\subsection{Optimization}
\label{method:optimization}
Once we fit the parameters for the VAE with the cost predictor, we can choose new designs to query by minimizing the predicted costs with $f_{\pi}$.
In our experiments, we instantiate $f_{\pi}$ with a feed-forward neural network and perform gradient descent directly in the latent space by differentiating through $f_\pi$ with respect to its inputs.

Naively optimizing the predicted cost without constraints yields poor results \cite{nguyen2016synthesizing,gomez2018automatic, griffiths2020constrained} because the cost predictor is only accurate near regions where training examples are available.
We propose \emph{prior regularization}, a means of softly constraining optimized latents to stay near the origin where the majority of the dataset lies.
We optimize latents according to a linear combination of predicted cost and prior log-probability:
\begin{equation}
\label{eq:latent_search}
    g(z) = f_{\pi}(z) - \gamma \log p_{\theta}(z)
\end{equation}
where $\gamma$ is a hyperparameter to control the strength of the prior regularization. 
While \cite{tripp2020sample} propose constraining search to a box around the origin for the same reason, we note that in high-dimensional space a box has exponentially many corners, and so a box large enough to contain most of the data mass is likely to have many uninhabited regions.

To balance quality and diversity of our samples, we also initialize the starting latent variables close to valid and high-performing circuits by a cost-weighted sampling method. 
Specifically, we sample circuits from the current dataset (Line 6 in Algorithm \ref{alg:circuitvae}) proportional to their datapoint weights (Equation \ref{eq:weighting_function}).
For a sample design $x$, we obtain an initial latent variable $z_0$ from the posterior $q_{\phi}(z|x)$ (Line 7) and the gradient descent on $g(z)$ starts at $z_0$. 
This procedure ensures that latents are initialized in high-probability and low-cost regions, while being diverse enough to provide good training data for the next round.

We perform gradient descent for a fixed number of steps and capture the latent variable values after every $t$ steps to get $z_t, z_{2t}$, etc. (Line 8). For each $z_t$, we decode to a distribution over $\X$ with $p_{\theta}(x|z_t)$ and sample a design $x$ to query its cost $c$ (Line 9 and 10). CircuitVAE (Algorithm \ref{alg:circuitvae}) repeats the training and optimization loop multiple times. In practice, we can parallelize latent space gradient descent (Line 8) to further accelerate the search. \autoref{fig:progression} presents an example optimization trajectory for CircuitVAE.

\section{Experiments}


\begin{figure*}[t]
    \centering
    \includegraphics[width=0.9\textwidth]{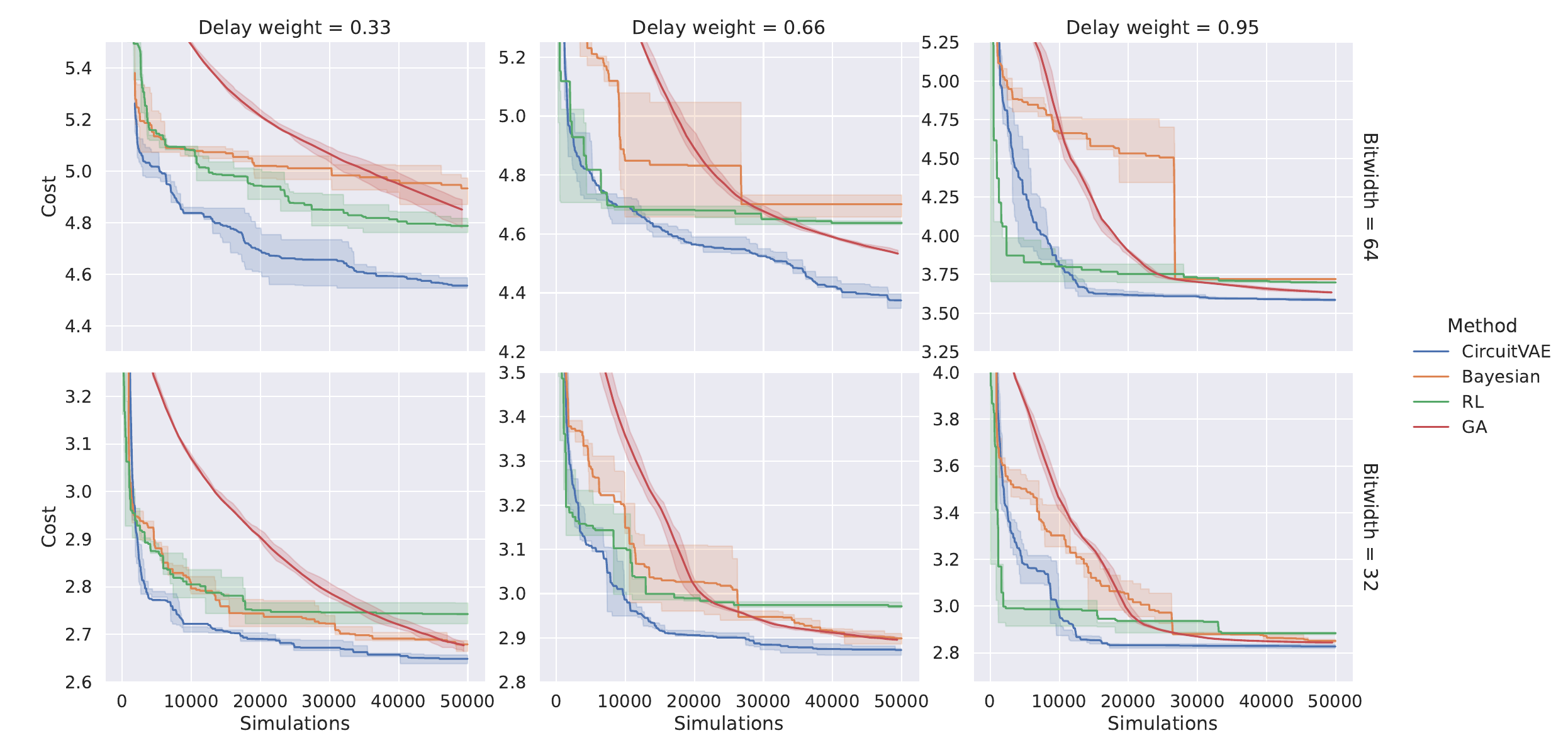}
    \caption{Curves of circuit cost (lower is better) vs simulation budget across a range of circuit sizes (rows) and timing constraints (columns). CircuitVAE consistently achieves lower costs at fewer simulations. The first and second row are 64-bit and 32-bit design tasks, respectively.}
\label{fig:cost_synths}
\vspace{-0.1in}
\end{figure*}

\subsection{Training and evaluation details}
We evaluated circuits following \cite{prefixrl}.
Prefix graphs generated by CircuitVAE are first legalized by inserting missing parents of existing nodes---this may be considered part of the objective function, so our cost predictor learns to infer the same value for equivalent circuits.
We then compile the legalized graph into a netlist using the 45-nanometer Nangate45 cell library \cite{ajayi2019openroad} and then physically synthesize the circuit using OpenPhySyn \cite{Agiza2020OpenPhySynAO}. $N$-bit prefix graphs are represented with in a $N \times N$ matrix as in \cite{prefixrl}.
Our encoder and decoder were each $\sim$ 1M-parameter CNNs autoencoding the computation graph with a 2-layer MLP as the cost predictor.
We represent circuits using the grid format described by \cite{prefixrl}.
All experiments used one A100 GPU and 24 CPU cores.

\subsection{Comparing search algorithms}
\label{subsec:comparison}

\begin{table}
  \caption{Detailed comparison of methods in the 64-bit, high-budget setting}
  \centering
  \hspace{-0.5em} 
  \resizebox{\columnwidth}{!}{%
  \begin{tabular}{lllllll}
    \toprule
    $\omega$ & Alg. & Cost & Area ($\text{µm}^2$) & Delay (ns) & VAE speedup \\
    \midrule
    0.33 & VAE & \textbf{4.54} (4.52 - 4.55) & \textbf{449} (446 - 452) & \textbf{0.465} (0.446 - 0.468) & -- \\
         & GA  & 4.65 (4.59 - 4.67) & 463 (458 - 465) & 0.480 (0.451 - 0.487) & 3.03 (2.15 - 3.45) \\
         & RL  & 4.72 (4.71 - 4.72) & 471 (464 - 482) & 0.473 (0.462 - 0.474) & 3.36 (2.46 - 3.48) \\
         & BO  & 4.94 (4.80 - 4.97) & 470 (468 - 475) & 0.536 (0.512 - 0.542) & 7.34 (5.99 - 8.50) \\
    \midrule
    0.66 & VAE & \textbf{4.31} (4.31 - 4.38) & \textbf{572} (572 - 579) & \textbf{0.360} (0.359 - 0.363) & -- \\
         & GA  & 4.44 (4.40 - 4.45) & 598 (587 - 606) & 0.363 (0.362 - 0.364) & 1.88 (1.65 - 1.93) \\
         & RL  & 4.63 (4.63 - 4.64) & 650 (650 - 656) & 0.368 (0.364 - 0.369) & 2.69 (2.18 - 4.88) \\
         & BO  & 4.66 (4.64 - 4.72) & 635 (607 - 654) & 0.388 (0.369 - 0.397) & 3.15 (1.22 - 10.36) \\
    \midrule
    0.95 & VAE & \textbf{3.58} (3.58 - 3.59) & 860 (834 - 902) & \textbf{0.333} (0.330 - 0.333) & -- \\
         & GA  & 3.60 (3.60 - 3.61) & 868 (855 - 872) & 0.334 (0.334 - 0.335) & 2.30 (2.21 - 5.49) \\
         & RL  & 3.70 (3.69 - 3.70) & \textbf{801} (800 - 803) & 0.347 (0.347 - 0.347) & 3.26 (3.18 - 5.82) \\
         & BO  & 3.72 (3.71 - 3.72) & 827 (806 - 836) & 0.347 (0.347 - 0.350) & 2.47 (1.96 - 2.48) \\
    \bottomrule
  \end{tabular}%
  }
  \label{table:methods}
  \vspace{-0.1in}
\end{table}

We compared CircuitVAE to three alternative search algorithms on the task of optimizing adders across a range of simulation budgets. Our primary baseline is PrefixRL (``RL''), the prior state-of-the-art reported by \cite{prefixrl}.
We also compared against a genetic algorithm (``GA'') directly optimizing a bitvector representation of the circuit; we used the first few generations of GA as the initial data to train CircuitVAE.
Finally, we compared against a variant of CircuitVAE which employs Bayesian optimization (BO) in the latent space, a practice which has become common \cite{tripp2020sample}.

Our results are summarized in \autoref{fig:cost_synths}: in all settings, CircuitVAE outperforms all other methods at every budget.
The 64-bit setting (the most difficult) is detailed in \autoref{table:methods}, which lists the cost, area, and delay of the best single adder found by each method in less than 70,000 simulations.
We also report the ``VAE speedup'', the simulation budget for each method to produce its best adder divided by the simulation budget for CircuitVAE to obtain an equivalent or better circuit.
In almost all cases, we find that CircuitVAE requires less than half the simulations to obtain equal performance, and in many cases less than one-third.


We repeated this experiment for bitwidths in \{32, 64\} and delay weights in \{0.33, 0.66, 0.95\}.
For CircuitVAE and Bayesian experiments, we launched runs with approximately 1,000, 5,000, 10,000, and 30,000 initial datapoints and grouped these runs into a single curve to report performance across a range of budgets; the initial simulations required to build the dataset were counted against these methods' budgets.
We ran each experiment with five different random seeds and independently collected initial datasets, and report the median and interquartile ranges across these runs.

We found gradient-based search outperforms latent Bayesian search in all settings.
We hypothesize that our higher-capacity neural cost predictor can learn more from large datasets than the Bayesian surrogate model, and that mitigating overoptimization with prior-regularized search enables quickly identifying promising candidates.

\subsection{Ablations}
\label{subsec:ablations}

\begin{figure}[t]
    \centering
    \includegraphics[width=0.45\textwidth]{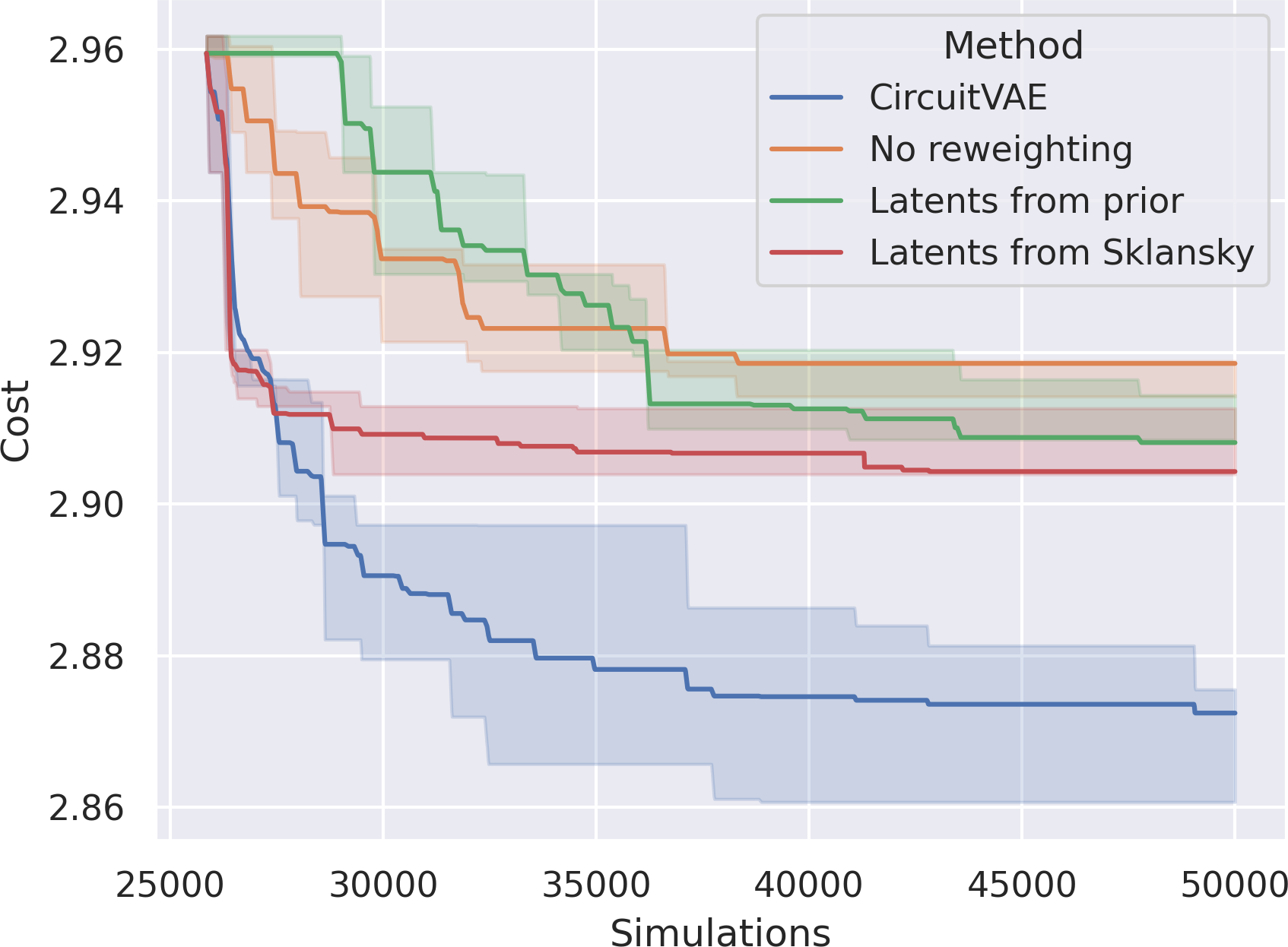}
    \caption{Ablating search and training methods.}
    \label{fig:abl_curves}
    \vspace{-0.1in}
\end{figure}

\begin{figure}[t]
    \centering
    \includegraphics[width=0.45\textwidth]{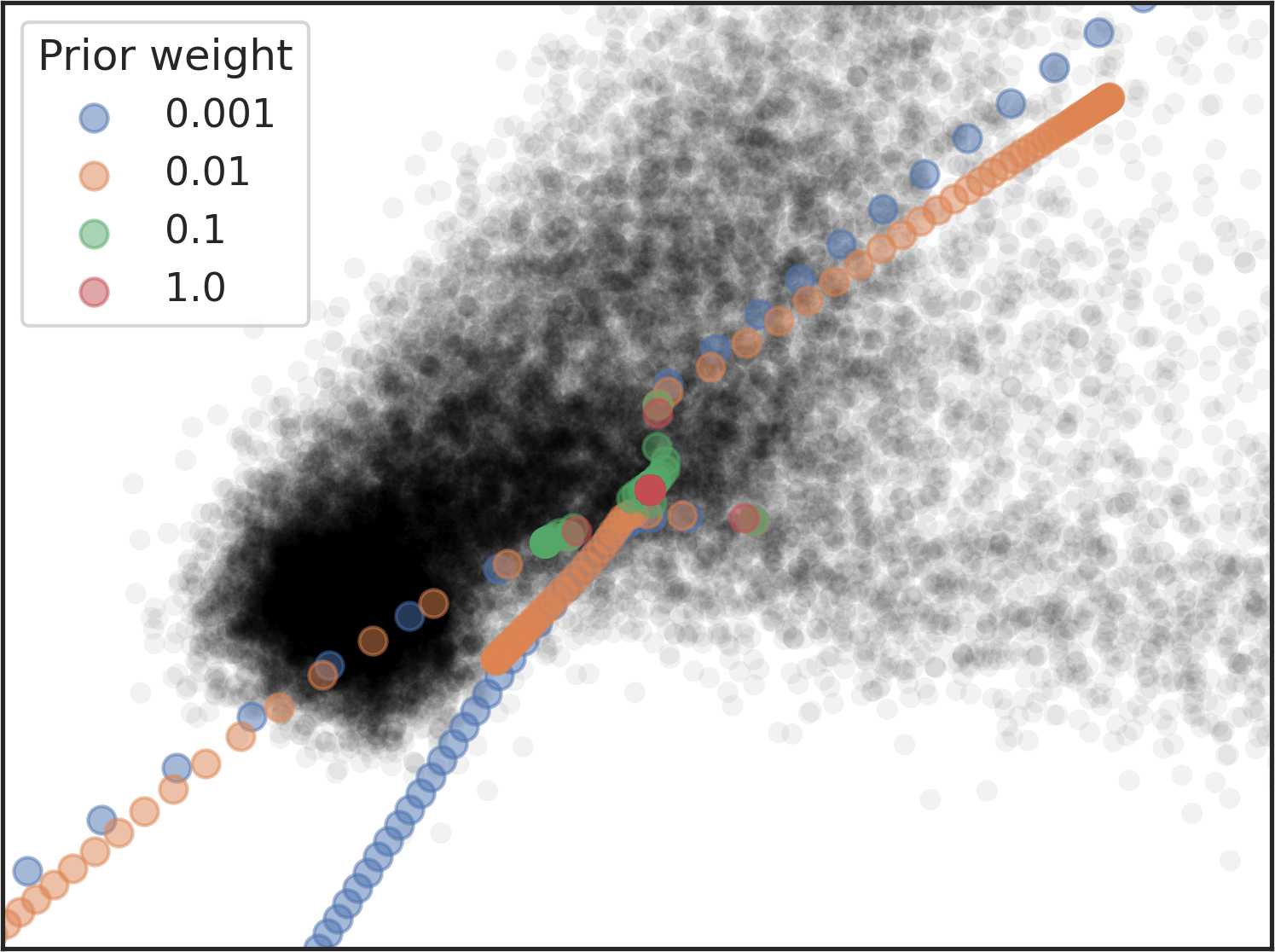}
    \caption{The effect of changing prior weight $\gamma$ on latent search trajectories. Low values of $\gamma$ result in trajectories ending far away from training data.}
    \label{fig:abl_priorweight}
    \vspace{-0.1in}
\end{figure}

We ablated each of the components of CircuitVAE to understand their individual contributions.
All of these experiments were conducted on 32-bit adders, with a delay weight of 0.66 and the largest initial dataset.
In \autoref{fig:abl_curves}, we tested:
\begin{itemize}
    \item Removing data reweighting \citep{tripp2020sample}, which leads training to get stuck when new datapoints have a negligible impact on the overall distribution.
    \item Replacing the cost-weighted latent distribution with the prior or the latent encoding of Sklansky. Starting the search from a good adder (Sklansky) outperforms sampling from the prior, but both underperform our adaptive initialization. 
\end{itemize}


In \autoref{fig:abl_priorweight}, we examined the ability to control search by controlling the prior regularization term $\gamma$.
At low values of $\gamma$ (blue and orange), latent trajectories quickly exit the region around the training data (gray) and overfit the cost predictor, yielding much higher costs than the model predicts.
At higher values (green and red), trajectories stay near the origin, which prevents overfitting but limits exploration and sample diversity.
We found the best results from sampling values of $\gamma$ per latent trajectory log-uniformly between 0.01 and 0.1, and used this setting for all other experiments.

\subsection{Designing real-world circuits}
\label{subsec:circuitE}

\begin{figure}[t]
    \centering
    \includegraphics[width=0.45\textwidth]{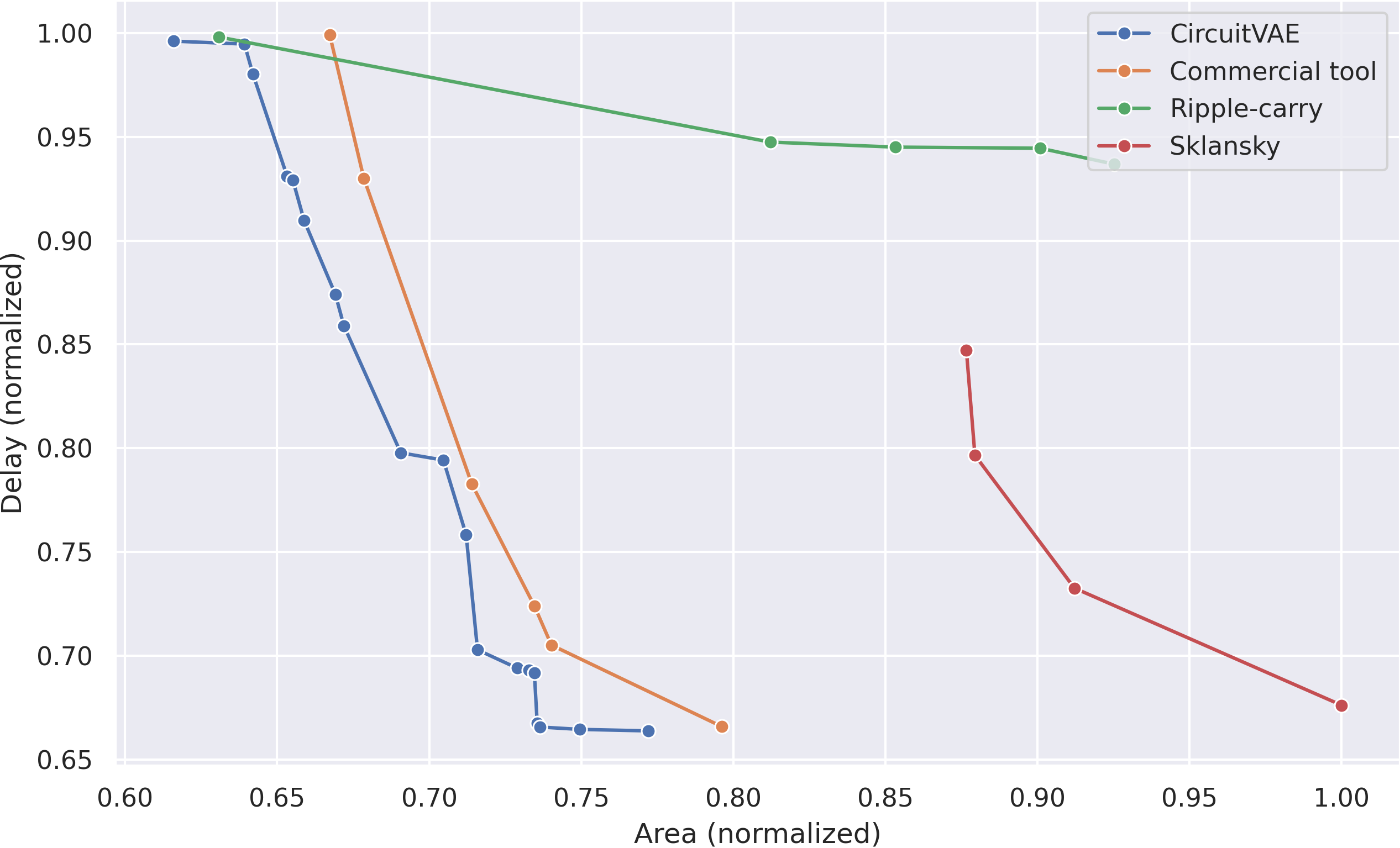}
    \caption{Comparing the area-delay Pareto frontiers of 8nm circuits in a realistic setting. CircuitVAE's designs Pareto-dominate both human-designed circuits and a commercial design tool.}
    \label{fig:circuitE}
    \vspace{-0.1in}
\end{figure}

To evaluate CircuitVAE in a more realistic setting, we tried using it in place of a commercial tool in a real-world datapath design.
We used CircuitVAE to design 31-bit adders at delay weights in \{0.3, 0.6, 0.95\} using OpenPhySyn as described above; however, we generated netlists with a proprietary 8nm cell library, and used bit input and output timings captured from a complete datapath.
Then, we synthesized the most promising designs using a commercial design tool.
Note the domain gap in the cost function between training and evaluation: the commercial tool makes different choices with respect to netlist buffering, gate sizing, cell placement, etc. 
Nevertheless, as \autoref{fig:circuitE} shows, we managed to Pareto-dominate both the design tool's provided adders and common human-designed adders.

\subsection{Designing gray-to-binary converts}
\label{subsec:recoder}

\begin{figure}[t]
    \centering
    \includegraphics[width=0.45\textwidth]{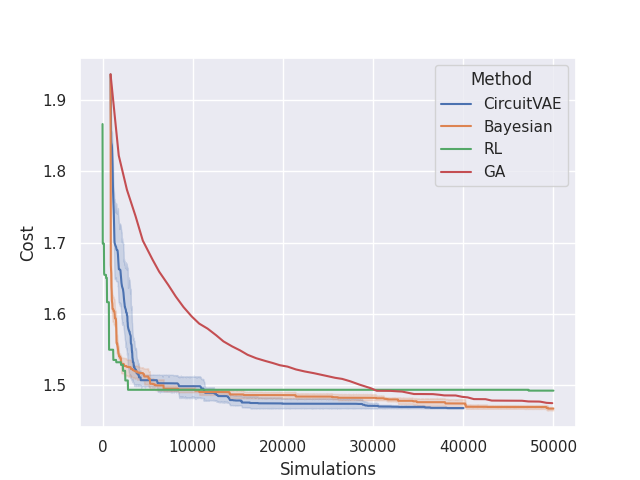}
    \caption{Curves of circuit cost (lower is better) vs simulation budget for gray-to-binary converters. CircuitVAE consistently achieves lower costs at fewer simulations.}
    \label{fig:recoder}
    \vspace{-0.1in}
\end{figure}

To showcase the generality of our CircuitVAE framework, we also designed a 26-bit gray-to-binary converter \cite{doran2007gray}. For this experiment, we target a delay weight of 0.6 and use the Nangate45 cell library to compile netlists. \autoref{fig:recoder} shows comparison results with the same set of baseline methods as in \autoref{subsec:comparison}. CircuitVAE outperforms all baselines as well in this design task.

\begin{figure}[t]
    \centering
    \includegraphics[width=0.4\textwidth]{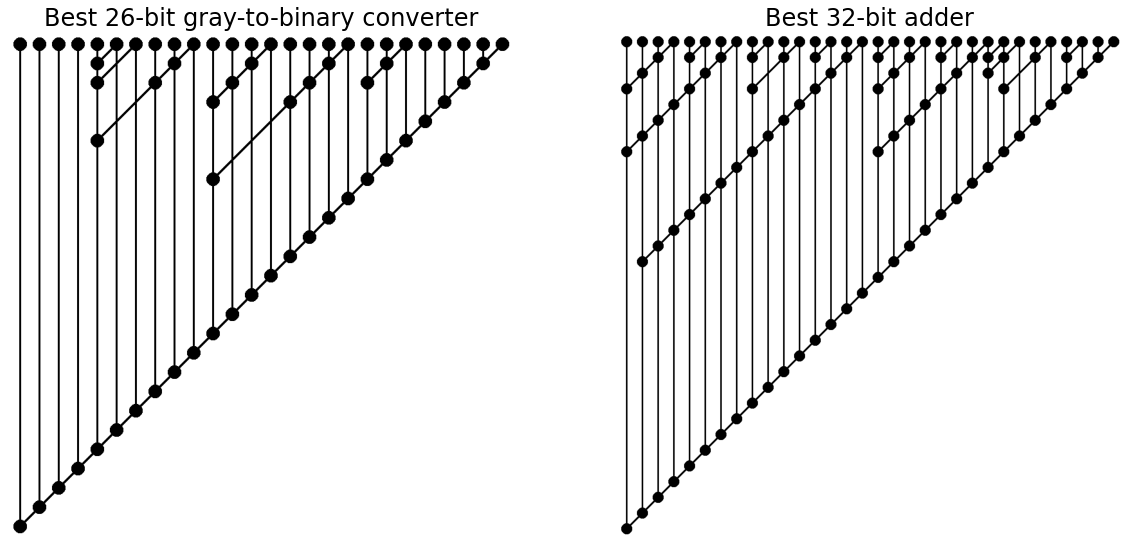}
    \caption{Best designs for gray-to-binary converter and adder. Their structural differences validate the two tasks are fundamentally different.}
    \label{fig:best_designs}
    \vspace{-0.1in}
\end{figure}

Finally, \autoref{fig:best_designs} shows the best designs for the gray-to-binary converter and a 32-bit adder. Although both design tasks targeted similar delay weights, there are significant structural differences in best designs discovered by CircuitVAE. They demonstrate that CircuitVAE can flexibily adapt to different circuit types.

\section{Conclusion}
In this work, we demonstrated that CircuitVAE can efficiently optimize binary adders and gray-to-binary converters, even in the difficult cases of large circuits and tight timing constraints.
However, the authors optimistically believe that the impact of this work extends these two classes of circuits.
Our method may be applied unchanged to optimize other prefix computations, such as leading zero detectors; by replacing the prefix graph with another data structure, one might also optimize multipliers or other circuits.
We hope to address these exciting possibilities in future work.

\renewcommand*{\bibfont}{\scriptsize}
{
\bibliographystyle{ACM-Reference-Format}
\bibliography{main}
}

\end{document}